\pdfoutput=1

\documentclass[11pt]{article}

\usepackage{emnlp2021}

\usepackage{times}
\usepackage{latexsym}

\usepackage[T1]{fontenc}

\usepackage[utf8]{inputenc}

\usepackage{microtype}

\def\tabref#1{Table~\ref{tab:#1}}

\def\tablabel#1{\label{tab:#1}\label{p:#1}}

\def\secref#1{\S\ref{sec:#1}}
\def\seclabel#1{\label{sec:#1}}

\newcounter{notecounter}

\usepackage{multirow}
\usepackage[caption=false]{subfig}
\usepackage{graphicx}
\usepackage{bm}
\usepackage{amssymb}
\usepackage{lscape}
\usepackage{hologo}

\usepackage{CJKutf8}

\title{
Discrete and Soft  Prompting for
Multilingual Models
}

\author{
  Mengjie Zhao {\normalfont and} Hinrich Sch\"{u}tze\\
  CIS, LMU Munich, Germany\\
  {\tt mzhao@cis.lmu.de}
}

\begin{document}
\maketitle

\begin{abstract}
It has been shown for English that discrete and soft prompting  perform strongly
in few-shot learning with pretrained language models (PLMs).
In this paper, we show that discrete and soft prompting perform better
than
finetuning in multilingual cases:
Crosslingual transfer
and
in-language training of
multilingual natural language inference.
For example, with 48 English training examples, finetuning obtains
33.74\% accuracy in crosslingual transfer, barely surpassing the
majority baseline (33.33\%).
In contrast, discrete and soft
prompting outperform finetuning, achieving 36.43\% and 38.79\%.
We also
demonstrate good performance of prompting
with training data in multiple languages other than English.

\end{abstract}

\section{Introduction}
\seclabel{intro}
Prompting strongly outperforms finetuning
\citep{devlin-etal-2019-bert} when adapting pretrained language
models (PLMs;
\citet{devlin-etal-2019-bert,conneau-etal-2020-unsupervised}) to
downstream  tasks in the low-resource regime
\citep{gpt3,schick2020s,gao2020making,tam2021improving,le2021many},
i.e., \emph{few-shot learning}, a more realistic scenario
than having tens of thousands of annotations, \emph{even for English}
\citep{yu-etal-2018-diverse,yin-etal-2020-universal,ram2021few}.

In contrast to finetuning, which learns discriminative classifiers
for tasks like natural language inference (NLI; \citet{10.1007/11736790_9,bowman-etal-2015-large}),
prompting reformulates the classification
task to generative
text-to-text \citep{t5paper} or
cloze-style \citep{McCann2018decaNLP,gpt3} queries which are
given to a PLM
to answer.
For example, the NLI task of
assigning 
premise ``They whinnied, eyes wide'' and 
hypothesis ``Their eyes were open wide'' to
class ``entailment'' can be reformulated as:\\[0.15cm]
\begin{small}
  \underline{\smash{They whinnied, eyes wide}} \texttt{.}
  \texttt{Question:} \underline{\smash{Their eyes}} \underline{\smash{were open wide}} \texttt{?} \texttt{Answer:} \underline{\ \ \ } \texttt{.}
\end{small}\\[0.15cm]
The PLM is requested to fill in, for the blank (\underline{\ \ \ }), the
word ``yes'', which is mapped to ``entailment''.

Prompting makes a human description of the task
available in learning. Also,
``filling in the blank'' is well aligned with the
pretraining objective (masked/autoregressive language modelling \citep{devlin-etal-2019-bert,radford2019language,xlnet}), likely
to deliver better performance in few-shot learning \citep{ram2021few}.

In this paper, we investigate
the effectiveness of prompting in
multilingual tasks, which -- 
despite the success of prompting in English --
is largely unexplored.
We address two main research questions:
(RQ1) Does the strong few-shot performance of
prompting transfer to other languages from English?
(RQ2) As the cost of few-shot
non-English annotations is affordable
\citep{garrette-baldridge-2013-learning,
  lauscher-etal-2020-zero,zhao-etal-2021-closer}, can
we directly prompt PLMs in languages other than English or
do we have to go through the (generally best resourced)
intermediary of English? 

In this work, 
we systematically compare
two popular prompting methods --
discrete and soft prompting --
with finetuning
in the few-shot multilingual  NLI task
and 
show that \emph{prompting is superior}:
(i) The strong few-shot learning performance
of prompting transfers to
other languages from English: It outperforms finetuning in
crosslingual transfer (RQ1; \secref{zstransfer}).
(ii) Directly querying the multilingual PLM with
few-shot non-English prompts  achieves competitive performance,
without relying on crosslingual
transfer from English (RQ2; \secref{inlanguageprompting}).

\section{Related Work}
GPT3 \citep{gpt3} succeeds in few-shot NLU
tasks with ``in-context learning'': A natural language prompt
describing the NLU task is prepended to an input example; GPT3 is
then capable of making accurate predictions \emph{without updating its
  parameters}.  However, the number of parameters in GPT3 is
prohibitively large (175B).

Integrating gradient descent into prompting, smaller (w.r.t. GPT3)
PLMs also achieve good few-shot performance.
Like GPT3, \textbf{discrete prompting}
uses natural language to describe NLU tasks.
\citet{schick2020s}, \citet{tam2021improving}, \citet{le2021many} use human-designed prompts.
\citet{gao2020making} leverage T5 \citep{t5paper} to generate prompts.
\citet{shin-etal-2020-autoprompt} use extra training data to search tokens
for constructing the prompts.
Discrete prompting naturally inherits interpretability from
the task descriptions.

\textbf{Soft prompting} relaxes the constraint that a prompt
needs to be composed of discrete tokens.
Instead, it learns the prompt in
the continuous space with SGD.
\citet{qin2021learning} and \citet{zhong2021factual} learn soft prompts
eliciting more knowledge \citep{petroni-etal-2019-language}
from PLMs than
discrete prompts.
Similar to soft prompting but with the PLM being frozen,
\citet{li2021prefix} propose
prefix-tuning
to encourage
PLMs to solve
generation tasks with
high parameter-efficiency \citep{houlsby2019parameter,zhao-etal-2020-masking}.
\citet{lester2021power} demonstrate that soft prompting benefits from
scaling up the number of PLM parameters.
\citet{liu2021gpt} show that GPT \citep{radford2019language} can
solve NLU tasks \citep{superglue} with soft prompting.

All of this work focuses on English.
We show that discrete and soft prompting
perform better than finetuning in
few-shot 
crosslingual natural language inference (XNLI; \citet{conneau2018xnli})
with multilingual PLMs (XLM-RoBERTa; \citet{conneau-etal-2020-unsupervised}).
We conduct experiments on NLI because
it is one of
the most representative and challenging
NLU tasks \citep{10.1007/11736790_9,bowman-etal-2015-large},
and has been commonly used in
prior work on prompting.

\section{Method}
\subsection{Finetuning}
We follow the standard finetuning method
\citep{devlin-etal-2019-bert}: A linear classifier layer is
initialized and stacked on top of the PLM; the whole model is
then trained on the few-shot NLI dataset (\secref{datasetandsetup}).

\subsection{Prompting}
\seclabel{enprompting}
\textbf{Discrete prompting (DP)}.
Following \citet{schick2020s}, \citet{le2021many}, we reformulate the 
NLI examples (cf.\ example in \secref{intro})
into cloze-style questions using a human-designed
prompt. Specifically, 
we ask
the PLM to fill in the blank (\underline{\ \ \ }) in sentence:\\[0.15cm]
\begin{small}
\underline{\smash{Premise}} \texttt{.} \texttt{Question:}
\underline{\smash{Hypothesis}} \texttt{?} \texttt{Answer:} \underline{\ \ \ } \texttt{.}
\end{small}\\[0.15cm]
\underline{\smash{Premise}} and \underline{\smash{Hypothesis}}
are a pair of sentences from the NLI dataset.
The gold labels are
mapped to words
in the PLM vocabulary.
Concretely, we use following mapping (\textbf{verbalizer};
\citet{schick2020s}): ``entailment''$\rightarrow$ ``yes'';
``contradiction''$\rightarrow$ ``no''; ``neutral''$\rightarrow$
``maybe''.
The optimization objective is to minimize the cross-entropy loss
between the predicted and the gold words representing the three
classes.

\textbf{Soft prompting} (\textbf{SP}; 
\citet{li2021prefix,qin2021learning,zhong2021factual,liu2021gpt})
leverages prompts containing 
``pseudo tokens'' that are not part of the PLM vocabulary.
In this work, we ask a PLM to fill in the blank
(\underline{\ \ \ }) in sentence:\\[0.15cm]
\begin{small}
\underline{\smash{Premise}} \texttt{.} \underline{\smash{Hypothesis}} \texttt{?}
<$v_1$><$v_2$><$v_3$><$v_4$> \underline{\ \ \ \ } \texttt{.}
\end{small}\\[0.15cm]
where each <$v_i$>, $i \in $ \{1, 2, 3, 4\} is associated with a randomly
initialized trainable vector (in the PLM's lowest embedding layer)
$\bm{v}_i \in \mathbb{R}^d$, where $d$ is the hidden dimension size of the
embedding layer.
Directly using $\bm{v}_i$ yields sub-optimal task performance:
\citet{li2021prefix} reparameterize $\bm{v}_i$ with another
trainable matrix and then feed it forward through an
MLP. Here, we adopt
\citet{liu2021gpt}'s approach. They feed [$\bm{v}_1, \bm{v}_2, \bm{v}_3, \bm{v}_4$]
through an
LSTM \citep{lstmpaper} and use the outputs.
PLM parameters, LSTM parameters, and $\bm{v}_i$ are jointly
trained.  Our SP and DP have the same training objective and verbalizer.

\textbf{Mixed prompting (MP)}. We also experiment with a
simple combination of DP and SP, by asking the PLM to
fill in the blank
(\underline{\ \ \ }) in sentence:\\[0.15cm]
\begin{small}
  \underline{\smash{Premise}} \texttt{.} \texttt{Question:} \underline{\smash{Hypothesis}} \texttt{?}
  <$v_1$><$v_2$><$v_3$><$v_4$> \texttt{Answer:} \underline{\ \ \ \ } \texttt{.}
\end{small}\\[0.15cm]
MP includes human descriptions of NLI as in DP and
learns ``soft prompts'' as in SP.

\begin{table*}[h!]
  \small\centering\renewcommand{\arraystretch}{1.15} \setlength{\tabcolsep}{2pt}
\begin{tabular}{l|l|l|l}
                                         &    & Prompt & Verbalizer           \\ \hline
\multicolumn{1}{c|}{\multirow{3}{*}{EN}} & DP & \underline{\smash{Premise}} \texttt{.} \texttt{Question:} \underline{\smash{Hypothesis}} \texttt{?} \texttt{Answer:} \underline{\ \ } \texttt{.} & Entailment $\rightarrow$ yes \\
\multicolumn{1}{c|}{}                    & SP & \underline{\smash{Premise}} \texttt{.} \underline{\smash{Hypothesis}} \texttt{?} <$v_1$>...<$v_4$> \underline{\ \ } \texttt{.}& Contradict $\rightarrow$  no  \\
\multicolumn{1}{c|}{}                    & MP & \underline{\smash{Premise}} \texttt{.} \texttt{Question:} \underline{\smash{Hypothesis}} \texttt{?} <$v_1$>...<$v_4$> \texttt{Answer:} \underline{\ \ } \texttt{.}& Neutral $\rightarrow$  maybe  \\ \hline
\multicolumn{1}{c|}{\multirow{3}{*}{TR}} & DP & \underline{\smash{Premise}} \texttt{.} \texttt{Soru:} \underline{\smash{Hypothesis}} \texttt{?} \texttt{Cevap:} \underline{\ \ } \texttt{.}  & Entailment $\rightarrow$ Evet \\
\multicolumn{1}{c|}{}                    & SP & \underline{\smash{Premise}} \texttt{.} \underline{\smash{Hypothesis}} \texttt{?} <$v_1$>...<$v_4$> \underline{\ \ } \texttt{.}& Contradict $\rightarrow$  hiçbir  \\
\multicolumn{1}{c|}{}                    & MP & \underline{\smash{Premise}} \texttt{.} \texttt{Soru:} \underline{\smash{Hypothesis}} \texttt{?} <$v_1$>...<$v_4$> \texttt{Cevap:} \underline{\ \ } \texttt{.}& Neutral $\rightarrow$  belki            \\ \hline
\end{tabular}
\caption{Prompts and verbalizers in English (EN) and
  Turkish (TR).  ``<$v_1$>...<$v_4$>''=``<$v_1$><$v_2$><$v_3$><$v_4$>''.
  Appendix \secref{translatedpattern}
  shows translated prompts/verbalizers of the languages used in our experiments. }
\tablabel{languagepatterns}
\end{table*}


\subsection{Non-English prompting}
\seclabel{translateprompt}
We also explore the results of prompting the PLM with languages other
than English, of which the \emph{few-shot} annotation cost is
affordable
\citep{garrette-baldridge-2013-learning,lauscher-etal-2020-zero, zhao-etal-2021-closer}.

For a non-English language $\mathcal{L}$ like Turkish, we
translate\footnote{
  We use Google Translate due to the simplicity of our prompt.
  Specialized bilingual dictionaries can also be used.
} the English
prompting words (\secref{enprompting}) ``Question'' and ``Answer''
into $\mathcal{L}$, e.g., ``Soru'' and ``Cevap'' in Turkish.
Correspondingly, the verbalizer maps gold labels into $\mathcal{L}$:
``entailment''$\rightarrow$ ``Evet'';
``contradiction''$\rightarrow$ ``hi\c{c}bir''; ``neutral''$\rightarrow$
``belki''.
\tabref{languagepatterns} presents
example prompts and verbalizers.

\section{Dataset and Setup}
\seclabel{datasetandsetup}
\textbf{Dataset}.
We conduct our experiments on natural language
inference datasets MNLI and XNLI
\citep{mnli,conneau2018xnli}.
MNLI provides multi-genre English NLI sentence pairs.
XNLI provides development and test splits of \emph{human-translated parallel} NLI
sentence pairs in 15 languages\footnote{
  The languages are English (EN), French (FR), Spanish (ES), German (DE), Greek
  (EL), Bulgarian (BG), Russian (RU), Turkish (TR), Arabic (AR),
  Vietnamese (VI), Thai (TH), Chinese (ZH), Hindi (HI), Swahili (SW),
  and Urdu (UR).
} and
the \emph{machine-translated}
MNLI training sets
in 14
languages.

For constructing the  \emph{few-shot training  set},
we randomly sample without
replacement K $\in$ \{1, 2, 4, 8, 16, 32, 64, 128, 256\} 
shots \emph{per class} from the EN MNLI training split.
Then we retrieve translations of this EN
training set
from XNLI to create the
few-shot training sets in the other
languages.

To simulate a realistic low-resource regime
\citep{kann-etal-2019-towards,perez2021true}, we use \emph{few-shot development sets}.
For EN, we sample the same number of shots (as training)
from the XNLI development split.  As a result, a 2-shot
experiment uses 2 training and 2 development shots per class.
For other languages, we retrieve the translations of the
English development set from XNLI.
Following \citet{conneau2018xnli}, we report \emph{accuracy} on XNLI
test.

\textbf{Setup}.
We conduct all experiments using the pretrained XLM-RoBERTa-base model
\citep{conneau-etal-2020-unsupervised}
containing 270M parameters
trained on 2.5 TB CommonCrawl data in 100 languages.
We use PyTorch \citep{pytorch} and
the HuggingFace framework \citep{wolf-etal-2020-transformers}.\footnote{Resources are available at \url{https://github.com/mprompting/xlmrprompt}}

We use batch size 32 for finetuning and 24 for prompting methods
due to resource limitations.
Following \citet{le2021many}, we use learning rate 1e-5 for both
finetuning and prompting.  Following the suggestions of
\citet{mosbach2021on}, \citet{zhang2021revisiting}, we train the model with a
large number of epochs (50) and select the checkpoint that
performs best on the development set.
We repeat each experiment 5 times
with different random seeds (\{1, 2, 3, 4, 5\}) and report  mean
and variance. Appendix \secref{checklist} shows our reproducibility checklist.

\begin{table*}[t]
\centering\scriptsize\setlength{\tabcolsep}{4pt}\renewcommand{\arraystretch}{1.05} 
\begin{tabular}{c|c|ccccccccccccccc|c}
Shots                & Method & AR          & BG            & DE            & EL            & \underline{\emph{EN}}            & ES            & FR            & HI            & RU            & SW & TH & TR & UR & VI & ZH               & $\overline{X}$ \\ \hline
-                    & MAJ & 33.33          &33.33          &33.33          &33.33          &33.33          &33.33          &33.33          &33.33          &33.33          &33.33          &33.33          &33.33          &33.33         &33.33         &33.33         &33.33 \\ \hline
\multirow{4}{*}{1}   & FT  & 32.53          &32.63          &32.94          &32.53          &32.91          &32.61          &32.65          &32.87          &32.67          &32.77          &33.11          &32.68          &32.87         &32.69         &32.77         &32.75 \\
                     & DP  & 32.08          &33.23          &32.97          &33.24          &33.15          &33.78          &34.08          &33.41          &33.78          &33.45          &33.00          &34.01          &31.99         &32.83         &33.64         &33.24\\
                     & SP  & \textbf{34.84} &\textbf{36.50} &\textbf{36.87} &\textbf{37.49} &\textbf{36.65} &\textbf{38.29} &\textbf{38.57} &\textbf{36.43} &\textbf{37.56} &\textbf{34.52} &\textbf{35.71} &\textbf{34.76} &\textbf{35.54}&\textbf{35.06}&\textbf{37.61}&\textbf{36.43}\\
                     & MP  & 32.31          &32.32          &33.03          &32.14          &33.29          &34.02          &33.74          &34.12          &33.03          &32.86          &32.18          &34.59          &32.65         &32.82         &33.35         &33.10\\\hline

\multirow{4}{*}{2}   & FT  & 33.16          &33.35          &33.82          &33.24          &33.43          &33.31          &33.30          &33.24          &33.29          &33.19          &33.40          &33.04          &33.20         &33.03         &33.29         &33.29\\
                     & DP  & 32.90          &35.11          &34.44          &34.69          &35.41          &35.43          &34.77          &34.11          &34.93          &32.97          &35.43          &\textbf{35.19} &32.75         &33.28         &36.46         &34.52\\
                     & SP  & \textbf{35.91} &\textbf{38.08} &\textbf{38.15} &\textbf{38.42} &\textbf{37.97} &\textbf{38.23} &\textbf{38.62} &\textbf{36.32} &\textbf{39.22} &\textbf{34.35} &\textbf{37.20} &34.75          &\textbf{35.52}&\textbf{36.67}&\textbf{37.71}&\textbf{37.14}\\
                     & MP  & 32.76          &34.25          &34.10          &33.26          &34.59          &33.81          &34.33          &33.75          &34.01          &33.88          &34.55          &34.51          &32.59         &33.83         &35.39         &33.97 \\\hline

\multirow{4}{*}{4}   & FT  & 33.86          &33.89          &33.73          &33.63          &33.90          &33.58          &33.55          &33.86          &33.58          &33.75          &33.71          &33.79          &33.67         &33.85         &33.78         &33.74 \\
                     & DP  & 35.42          &37.64          &38.85          &37.67          &39.50          &38.91          &38.26          &36.43          &37.54          &34.72          &37.76          &\textbf{37.23} &35.92         &36.02         &38.74         &37.37\\
                     & SP  & \textbf{38.04} &\textbf{40.46} &\textbf{40.08} &\textbf{40.79} &\textbf{41.84} &\textbf{39.78} &\textbf{41.10} &\textbf{37.55} &\textbf{41.72} &\textbf{35.81} &\textbf{39.23} &35.88          &\textbf{37.66}&\textbf{37.86}&\textbf{39.48}&\textbf{39.15}\\
                     & MP  & 33.14          &33.79          &35.16          &33.95          &36.26          &35.52          &35.44          &34.63          &34.21          &33.53          &35.96          &35.62          &33.51         &34.06         &37.10         &34.79\\\hline

\multirow{4}{*}{8}   & FT  & 32.85          &32.75          &33.05          &32.59          &33.06          &32.58          &32.80          &32.89          &32.88          &32.75          &33.14          &32.69          &33.05         &32.83         &32.65         &32.84 \\
                     & DP  & 32.73          &34.78          &34.79          &34.82          &36.39          &34.97          &35.17          &33.00          &34.59          &32.91          &35.14          &34.13          &33.14         &33.66         &35.56         &34.39\\
                     & SP  & \textbf{36.30} &\textbf{38.84} &\textbf{38.22} &\textbf{38.68} &\textbf{39.02} &\textbf{38.16} &\textbf{38.82} &\textbf{35.86} &\textbf{39.73} &\textbf{34.50} &\textbf{37.90} &35.11          &\textbf{35.61}&\textbf{37.41}&\textbf{37.17}&\textbf{37.42}\\
                     & MP  & 32.67          &33.24          &34.81          &33.18          &34.78          &34.66          &34.77          &34.76          &33.81          &33.07          &34.46          &\textbf{35.12} &32.69         &33.57         &36.34         &34.13 \\\hline

\multirow{4}{*}{16}  & FT  & 33.72          &34.09          &34.28          &33.49          &34.73          &33.82          &33.81          &33.08          &34.06          &33.69          &33.06          &33.57          &33.22         &34.01         &33.46         &33.74 \\
                     & DP  & 35.07          &37.07          &37.51          &37.43          &38.24          &36.91          &36.61          &35.85          &36.51          &33.84          &37.21          &35.74          &34.86         &35.77         &\textbf{37.86}&36.43\\
                     & SP  & \textbf{38.88} &\textbf{40.60} &\textbf{40.21} &\textbf{40.44} &\textbf{39.45} &\textbf{39.37} &\textbf{40.90} &\textbf{36.86} &\textbf{40.61} &\textbf{37.11} &\textbf{39.45} &\textbf{36.26} &\textbf{35.88}&\textbf{38.46}&37.35         &\textbf{38.79}\\
                     & MP  & 32.46          &33.02          &33.98          &32.59          &33.20          &34.54          &34.39          &34.30          &33.90          &33.28          &33.47          &34.69          &32.67         &33.28         &35.68         &33.70\\\hline

\multirow{4}{*}{32}  & FT  & 35.84          &36.28          &36.00          &36.11          &36.64          &36.02          &36.47          &35.41          &35.68          &35.33          &35.71          &35.90          &34.81         &36.10         &36.20         &35.90    \\
                     & DP  & \textbf{41.80} &\textbf{43.51} &\textbf{43.49} &\textbf{42.50} &43.65          &\textbf{42.83} &43.90          &39.30          &42.39          &37.51          &\textbf{40.51} &\textbf{42.01} &39.77         &41.91         &39.94         &\textbf{41.67}\\
                     & SP  & 40.30          &43.38          &42.08          &42.27          &44.72          &42.32          &42.34          &38.91          &\textbf{43.76} &\textbf{37.54} &39.97          &38.79          &38.83         &\textbf{42.09}&39.56         &41.12\\
                     & MP  & 40.95          &42.16          &42.61          &42.31          &\textbf{45.52} &41.22          &\textbf{44.67} &\textbf{40.17} &42.18          &36.52          &40.16          &41.21          &\textbf{40.48}&41.74         &\textbf{40.89}&41.52\\\hline

\multirow{4}{*}{64}  & FT  & 40.16          &39.56          &40.10          &39.87          &41.68          &40.34          &39.47          &39.53          &38.34          &39.64          &39.18          &39.50          &39.23         &40.85         &39.63         &39.81 \\
                     & DP  & \textbf{45.64} &\textbf{47.64} &48.05          &\textbf{46.94} &48.89          &44.95          &47.97          &\textbf{41.61} &\textbf{44.85} &\textbf{40.98} &\textbf{45.65} &\textbf{45.67} &43.37         &47.30         &\textbf{45.24}&\textbf{45.65}\\
                     & SP  & 43.48          &43.81          &45.99          &43.70          &49.04          &45.79          &46.11          &40.86          &44.51          &40.49          &44.68          &41.91          &40.09         &45.25         &44.17         &43.99\\
                     & MP  & 43.86          &46.01          &\textbf{48.22} &46.79          &\textbf{51.84} &\textbf{46.61} &\textbf{48.31} &40.11          &44.75          &37.84          &45.01          &44.82          &\textbf{43.95}&\textbf{48.28}&43.03         &45.30\\\hline

\multirow{4}{*}{128} & FT  & 43.50          &45.52          &45.60          &44.38          &46.94          &45.75          &46.00          &42.96          &44.94          &41.43          &43.27          &43.67          &41.78         &44.81         &44.79         &44.36 \\
                     & DP  & 46.23          &\textbf{50.49} &\textbf{50.99} &47.39          &\textbf{53.68} &48.53          &49.28          &44.77          &\textbf{46.93} &42.03          &47.95          &\textbf{49.56} &\textbf{44.21}&\textbf{48.92}&49.56         &48.03\\
                     & SP  & 44.78          &46.24          &45.30          &46.31          &49.45          &45.80          &46.37          &43.29          &44.95          &41.21          &45.64          &41.93          &41.18         &44.99         &45.73         &44.88\\
                     & MP  & \textbf{46.48} &47.98          &49.04          &\textbf{49.09} &52.55          &\textbf{49.66} &\textbf{50.34} &\textbf{47.03} &46.40          &\textbf{42.89} &\textbf{48.08} &48.45          &44.04         &48.15         &\textbf{50.47}&\textbf{48.04}\\\hline

\multirow{3}{*}{256} & FT  & 52.13          &54.57          &54.43          &54.00          &57.79          &\textbf{55.89} &55.39          &\textbf{50.65} &52.90          &\textbf{50.00} &51.22          &\textbf{52.31} &\textbf{48.57}&54.16         &52.10         &53.07\\
                     & DP  & \textbf{53.23} &55.59          &\textbf{55.39} &\textbf{55.05} &60.14          &50.64          &54.43          &46.10          &51.35          &45.26          &53.42          &50.83          &48.42         &\textbf{55.14}&52.72         &52.51\\
                     & SP  & 52.26          &\textbf{56.04} &53.02          &53.12          &\textbf{60.58} &54.80          &\textbf{55.79} &49.43          &52.49          &47.33          &\textbf{54.52} &52.08          &48.48         &54.54         &\textbf{54.59}&\textbf{53.27}\\
                     & MP  & 52.77          &53.98          &50.71          &54.63          &60.13          &51.64          &55.32          &49.58          &\textbf{53.50} &45.27          &53.37          &51.28          &47.16         &52.34         &53.80         &52.37 \\\hline
\end{tabular}
\caption{Zero-shot crosslingual transfer results in accuracy
  (\%). Each number is the mean performance of 5 runs, when using finetuning (FT),
  discrete prompting (DP), soft prompting (SP), and mixed prompting (MP). ``MAJ'':
  majority baseline; $\overline{X}$: macro average across 15
  languages. Please see  Appendix \tabref{zeroshottablewithvariance} for variances.}
\tablabel{zeroshottable}
\end{table*}


\section{Experiments}
\subsection{Zero-shot crosslingual transfer}
\seclabel{zstransfer}
We first compare prompting with finetuning in
\emph{zero-shot crosslingual transfer} \citep{
  pires-etal-2019-multilingual,
  conneau-etal-2020-unsupervised,
  artetxe-schwenk-2019-massively,hu2020xtreme}:
The PLM is trained on the EN few-shot dataset and
then directly evaluated on the test set of all
languages. \tabref{zeroshottable} reports the results.

\textbf{EN results}.
From column \underline{\emph{EN}} we observe that:
(i) As expected, all four methods benefit from more shots.
(ii) Prompting methods (DP/SP/MP) clearly outperform finetuning
especially in low-resource regimes.
For example, in the 4-shot experiment, SP
outperforms finetuning by $\approx$8 (41.84-33.90) accuracy points.
\tabref{qualitativetabel} displays some examples for which
SP outperforms finetuning.
The improvements become less significant when more shots are available, e.g., 256.
(iii) SP outperforms DP for most choices of shots (except
128),
evidencing the strength
of relaxing the ``discrete token'' constraint in DP
\citep{liu2021gpt,qin2021learning,zhong2021factual}. But we
give up 
the interpretability of DP for this better performance.
(iv) Performance of MP -- the combination of DP and SP -- is
decent, but not stellar.
Future work may explore advanced prompting
methods succeeding in
both task performance and interpretability.
We focus on DP and SP in following experiments.

\begin{table}[t]
\centering\small\setlength{\tabcolsep}{3pt}\renewcommand{\arraystretch}{1.2} 
\begin{tabular}{l|l}
Premise/Hypothesis                   & Prediction                        \\ \hline
This was the temper of the times.    & \multirow{2}{*}{``no'' (Contradict)}  \\
This wasn't the temper of the times. &                                   \\ \hline
We would go in there.                & \multirow{2}{*}{``maybe'' (Neutral)}  \\
We would enter there at 8pm.         &                                   \\ \hline
I hope to hear from you soon.        & \multirow{2}{*}{``yes'' (Entailment)} \\
I hope we talk soon.                 &                                             \\ \hline
\end{tabular}
\caption{Qualitative examples for which prompting outperforms finetuning.}
\tablabel{qualitativetabel}
\end{table}

\textbf{Crosslingual transfer results} closely follow
the trends of EN results: Prompting outperforms finetuning
when looking at the macro average $\overline{X}$.
One intriguing
finding is that DP successfully transfers the learned knowledge
to target languages, better than SP in some languages, using
the \emph{code-switched} prompt:
``
\underline{\smash{Premise}} \texttt{.} \texttt{Question:}
\underline{\smash{Hypothesis}} \texttt{?} \texttt{Answer:} \underline{\ \ \ } \texttt{.}
'' where \underline{\smash{Premise}} and \underline{\smash{Hypothesis}} are
non-English. Thus,
DP is able to leverage
the strong crosslingual ability of the multilingual PLM.
Like finetuning, prompting does not uniformly benefit the 14 non-English
languages. For example, the crosslingual transfer performance of HI/SW/UR is notably
inferior compared with other languages.

Overall, prompting outperforms finetuning in zero-shot
crosslingual transfer of NLI in the low-resource regimes.

\begin{table}[t]
\centering\scriptsize\setlength{\tabcolsep}{4pt}
\begin{tabular}{c|c|cccc}
Shots                & Method  & TR                & UR                & SW               & ZH     \\ \hline
\multirow{3}{*}{8}   & FT      &32.71              &32.83              & 32.80            &33.31   \\
                     & DP      &\textbf{38.02}     &\textbf{39.33}     & \textbf{33.84}   &\textbf{37.46}  \\
                     & SP      &35.41              &34.59              & 33.47            &34.39   \\\hline

\multirow{3}{*}{16}  & FT     &33.00              &33.78              & 33.46            &33.56   \\
                     & DP     &39.39              &\textbf{40.58}     & \textbf{34.48}   &\textbf{42.24}  \\
                     & SP     &\textbf{40.22}     &35.47              & 33.99            &35.64   \\\hline

\multirow{3}{*}{32}  & FT     &37.15              &34.23              & 34.52            &35.38   \\
                     & DP     &\textbf{48.79}     &\textbf{41.67}     & \textbf{37.52}   &\textbf{38.04}   \\
                     & SP     &43.62              &39.18              & 36.00            &35.13   \\\hline

\multirow{3}{*}{64}  & FT     &38.87              &35.90              & 36.37            &42.14   \\
                     & DP     &\textbf{48.97}     &\textbf{42.34}     & 37.72            &\textbf{44.73}   \\
                     & SP     &47.26              &39.12              & \textbf{37.98}   &40.75   \\\hline

\multirow{3}{*}{128} & FT     &40.84              &36.23              & 36.81            &43.16   \\
                     & DP     &\textbf{49.73}     &\textbf{45.22}     & \textbf{41.26}   &\textbf{49.24}   \\
                     & SP     &47.68              &40.96              & 40.42            &47.17   \\\hline

\multirow{3}{*}{256} & FT     &49.41              &40.12              & 42.17            &48.98   \\
                     & DP     &\textbf{52.61}     &\textbf{46.10}     & \textbf{47.69}   &\textbf{53.11}   \\
                     & SP     &51.21              &44.60              & 46.89            &52.76   \\\hline
\end{tabular}
\caption{In-language results in accuracy (\%). 
  Prompting (DP/SP) outperforms finetuning (FT). Please see Appendix \tabref{inlanguagetablewithvar} for variances.}
\tablabel{inlanguagetable}
\end{table}


\subsection{In-language prompting}
\seclabel{inlanguageprompting}
We next compare prompting with finetuning when
using non-English few-shot datasets.
Taking Turkish as an example, recall that we can use the Turkish
prompts (\secref{translateprompt}) and few-shot datasets from XNLI
(\secref{datasetandsetup}) to finetune/prompt the PLM
directly.

\tabref{inlanguagetable} shows results of in-language experiments
of Turkish, Urdu, Swahili, and Chinese.
We make two main observations: (i) Prompting still outperforms finetuning,
though the non-English prompts and verbalizers
are translated from EN simply using Google Translate.
(ii) In-language results are slightly worse
but competitive to transfer learning results
(\tabref{zeroshottable}).
We conjecture that two factors result in
the second observation.
First, some languages have a small amount of pretraining data.  For
example, Swahili has 1.6GB pretraining data
while English has 300GB
\citep{conneau-etal-2020-unsupervised}.
Thus, the PLM may not be well
pretrained for solving tasks in Swahili directly.
Second, the few-shot training data for non-English languages
is machine-translated (\secref{datasetandsetup}).
With better \emph{few-shot}
translations and in-language expertise, prompting possibly
could achieve even better results.

Overall, the experimental results show that
directly prompting PLMs with non-English languages is also an
effective way of solving NLU tasks in low-resource regimes.

\section{Conclusion}
We showed that prompting performs better than finetuning
in few-shot crosslingual transfer and in-language training
of multilingual natural language inference.
We hope our results will encourage more research about prompting
multilingual tasks and models.

Future work may explore using text-to-text models like T5 \citep{t5paper} 
or other autoregressive PLMs \citep{lewis-etal-2020-bart}.
Investigating PLMs containing even more parameters
is also promising as shown by
\citet{lester2021power}.

\section*{Acknowledgements}
We thank  
the anonymous reviewers for the insightful
comments and suggestions.
This work was funded by the European Research Council
(ERC \#740516).

\bibliography{custom}
\bibliographystyle{acl_natbib}

\clearpage
\appendix   
\section{Reproducibility Checklist}
\seclabel{checklist}
\subsection{Model architecture and number of parameters}
We use the \texttt{xlm-roberta-base}
model \citep{conneau-etal-2020-unsupervised}.
It contains 12 Transformer blocks with 768 hidden dimensions.  Each
block has 12 attention heads. Vocabulary size is 250K.  Overall the
model has 270M parameters and was pretrained on the on 2.5 TB of newly
created clean CommonCrawl data in 100 languages.

Following \citet{liu2021gpt}, we employ a bidirectional LSTM
in SP and MP. The hidden dimension is also 768 so the number of
LSTM parameters is $2 \times 4\times(768\times768+768\times768+768) \approx$10M.
Another MLP is used to project the concatination of LSTM states back
to 768-dimension which has $(768\times2\times768+768) \approx$1.2M. SP
and MP also have four learnable vectors $\bm{v}_i$ resulting in
$768\times4=3072$ parameters.

\subsection{Computing infrastructure}
All experiments are conducted on GeForce GTX 1080Ti.
For finetuning, we use batch size 32 and 4 GPUs.
Because prompting uses the masked language model objective
so we use a maximum batch size 24.
A single GPU is used for 1-shot experiments.
Two and three GPUs are used for 2- and 4-shot experiments.
Other experiments use 6 GPUs.

\subsection{Evaluation metrics}
Our code is available at \url{https://github.com/mprompting/xlmrprompt}.
We use the standard evaluation metric accuracy as \citet{conneau2018xnli}.
For finetuning, evaluation script path is \url{./finetuning/utils/eval_meters.py}.
For DP, evaluation script path is \url{./pet/pet/trainers/meters.py}.
For SP/MP, evaluation script path is \url{./sptuning/pet/trainers/meters.py}.

\subsection{Hyperparameter search}
We use the same learning rate (1e-5) as \citet{le2021many} who compare
prompting and finetuning in English NLU tasks. No learning rate
scheduling is used for clear comparisons.  For both finetuning and
prompting, the model is trained for 50 epochs and the checkpoint that
performs best on development set is selected for performance
evaluation.

\subsection{Datasets and preprocessing}
We retrieve the MNLI and XNLI datasets from the official websites:
\url{cims.nyu.edu/~sbowman/multinli} and
\url{cims.nyu.edu/~sbowman/xnli}.
We use the tokenizer in the HuggingFace framework
\citep{wolf-etal-2020-transformers} to preprocess the texts.  In all
experiments, the max sequence length is 256.

\section{Translated Prompts}
\seclabel{translatedpattern}
\begin{table}[t]
  \tiny\centering\renewcommand{\arraystretch}{1.2} \setlength{\tabcolsep}{3pt}
\begin{tabular}{l|l|l|l}
                                         &    & Prompt & Verbalizer           \\ \hline
\multicolumn{1}{c|}{\multirow{3}{*}{TR}} & DP & \underline{\smash{P}} \texttt{.} \texttt{Soru:} \underline{\smash{H}} \texttt{?} \texttt{Cevap:} \underline{\ \ } \texttt{.}  & Entailment $\rightarrow$ Evet \\
\multicolumn{1}{c|}{}                    & SP & \underline{\smash{P}} \texttt{.} \underline{\smash{H}} \texttt{?} <$v_1$>...<$v_4$> \underline{\ \ } \texttt{.}& Contradict $\rightarrow$  hiçbir  \\
\multicolumn{1}{c|}{}                    & MP & \underline{\smash{P}} \texttt{.} \texttt{Soru:} \underline{\smash{H}} \texttt{?} <$v_1$>...<$v_4$> \texttt{Cevap:} \underline{\ \ } \texttt{.}& Neutral $\rightarrow$  belki \\\hline
\multicolumn{1}{c|}{\multirow{3}{*}{SW}} & DP & \underline{\smash{P}} \texttt{.} \texttt{Swali:} \underline{\smash{H}} \texttt{?} \texttt{Jibu:} \underline{\ \ } \texttt{.}  & Entailment $\rightarrow$ ndio \\
\multicolumn{1}{c|}{}                    & SP & \underline{\smash{P}} \texttt{.} \underline{\smash{H}} \texttt{?} <$v_1$>...<$v_4$> \underline{\ \ } \texttt{.}& Contradict $\rightarrow$  hasi  \\
\multicolumn{1}{c|}{}                    & MP & \underline{\smash{P}} \texttt{.} \texttt{Swali:} \underline{\smash{H}} \texttt{?} <$v_1$>...<$v_4$> \texttt{Jibu:} \underline{\ \ } \texttt{.}& Neutral $\rightarrow$  labda \\\hline
\multicolumn{1}{c|}{\multirow{3}{*}{ZH}} & DP & \underline{\smash{P}} \texttt{.} \begin{CJK*}{UTF8}{gbsn}问题：\end{CJK*} \underline{\smash{H}} \texttt{?} \begin{CJK*}{UTF8}{gbsn}答案：\end{CJK*} \underline{\ \ } \texttt{.}  & Entailment $\rightarrow$ \begin{CJK*}{UTF8}{gbsn}是\end{CJK*} \\
\multicolumn{1}{c|}{}                    & SP & \underline{\smash{P}} \texttt{.} \underline{\smash{H}} \texttt{?} <$v_1$>...<$v_4$> \underline{\ \ } \texttt{.}& Contradict $\rightarrow$  \begin{CJK*}{UTF8}{gbsn}否\end{CJK*}  \\
\multicolumn{1}{c|}{}                    & MP & \underline{\smash{P}} \texttt{.} \begin{CJK*}{UTF8}{gbsn}问题：\end{CJK*} \underline{\smash{H}} \texttt{?} <$v_1$>...<$v_4$>  \begin{CJK*}{UTF8}{gbsn}答案：\end{CJK*} \underline{\ \ } \texttt{.}& Neutral $\rightarrow$  \begin{CJK*}{UTF8}{gbsn}也许\end{CJK*}

\end{tabular}
\caption{Prompts and verbalizers in Turkish (TR), Swahili (SW), and Chinese (ZH). }
\tablabel{all_languagepatterns}
\end{table}

\tabref{all_languagepatterns} shows the prompts and verbalizers used
in in-language experiments. We use Google Translate but more
specialized bilingual dictionaries can also be used.  For Urdu, we
show the prompt and verbalizer in the code repository.

\begin{table}[t]
\centering\scriptsize\setlength{\tabcolsep}{4pt}\renewcommand{\arraystretch}{1.2}
\begin{tabular}{c|c|cccc}
Shots                & Method  & TR               & UR               & SW               & ZH     \\ \hline
\multirow{3}{*}{8}   & FT      &32.71$\pm$0.61    &32.83$\pm$0.29    & 32.80$\pm$0.56   &33.31$\pm$0.27   \\
                     & DP      &38.02$\pm$1.14    &39.33$\pm$0.58    & 33.84$\pm$0.44   &37.46$\pm$0.62  \\
                     & SP      &35.41$\pm$0.30    &34.59$\pm$0.26    & 33.47$\pm$0.34   &34.39$\pm$0.25   \\\hline

\multirow{3}{*}{16}  & FT     &33.00$\pm$0.93     &33.78$\pm$0.58     & 33.46$\pm$0.91   &33.56$\pm$0.50   \\
                     & DP     &39.39$\pm$0.81     &40.58$\pm$0.67     & 34.48$\pm$0.86   &42.24$\pm$2.66  \\
                     & SP     &40.22$\pm$0.50     &35.47$\pm$0.61     & 33.99$\pm$0.17   &35.64$\pm$1.03   \\\hline

\multirow{3}{*}{32}  & FT     &37.15$\pm$1.78     &34.23$\pm$0.85     & 34.52$\pm$1.20   &35.38$\pm$0.47   \\
                     & DP     &48.79$\pm$0.40     &41.67$\pm$1.39     & 37.52$\pm$1.08   &38.04$\pm$1.00   \\
                     & SP     &43.62$\pm$0.67     &39.18$\pm$1.09     & 36.00$\pm$1.23   &35.13$\pm$0.75   \\\hline

\multirow{3}{*}{64}  & FT     &38.87$\pm$0.99     &35.90$\pm$1.26     & 36.37$\pm$1.13   &42.14$\pm$1.22   \\
                     & DP     &48.97$\pm$0.56     &42.34$\pm$0.91     & 37.72$\pm$0.81   &44.73$\pm$0.86   \\
                     & SP     &47.26$\pm$0.77     &39.12$\pm$1.36     & 37.98$\pm$1.48   &40.75$\pm$1.90   \\\hline

\multirow{3}{*}{128} & FT     &40.84$\pm$1.45     &36.23$\pm$0.19     & 36.81$\pm$1.85   &43.16$\pm$0.95   \\
                     & DP     &49.73$\pm$0.73     &45.22$\pm$0.57     & 41.26$\pm$1.48   &49.24$\pm$0.89   \\
                     & SP     &47.68$\pm$0.68     &40.96$\pm$1.23     & 40.42$\pm$1.53   &47.17$\pm$0.54   \\\hline

\multirow{3}{*}{256} & FT     &49.41$\pm$2.03     &40.12$\pm$1.77    & 42.17$\pm$1.77   &48.98$\pm$3.12   \\
                     & DP     &52.61$\pm$1.34     &46.10$\pm$0.40    & 47.69$\pm$1.12   &53.11$\pm$0.61   \\
                     & SP     &51.21$\pm$0.30     &44.60$\pm$0.91    & 46.89$\pm$0.81   &52.76$\pm$0.29   \\\hline
\end{tabular}
\caption{In-language results in accuracy (\%). 
  Prompting (DP/SP) outperforms finetuning (FT). We report mean and variance of 5 runs.}
\tablabel{inlanguagetablewithvar}
\end{table}


\section{More Results}
\tabref{inlanguagetablewithvar} and \tabref{zeroshottablewithvariance}
show performances with variances.

\clearpage
\begin{landscape}
\begin{table}[t]
\centering\scriptsize\setlength{\tabcolsep}{3pt}\renewcommand{\arraystretch}{1.1} 
\begin{tabular}{c|c|ccccccccccccccc|c}
Shots                & Method & AR          & BG            & DE            & EL            & \underline{\emph{EN}}            & ES            & FR            & HI            & RU            & SW & TH & TR & UR & VI & ZH               & $\overline{X}$ \\ \hline
-                    & MAJ & 33.33          &33.33          &33.33          &33.33          &33.33          &33.33          &33.33          &33.33          &33.33          &33.33          &33.33          &33.33          &33.33         &33.33         &33.33         &33.33 \\ \hline
\multirow{4}{*}{1}   & FT  & 32.53$\pm$1.08 &32.63$\pm$1.02 &32.94$\pm$0.50 &32.53$\pm$1.10 &32.91$\pm$0.62 &32.61$\pm$0.89 &32.65$\pm$0.95 &32.87$\pm$0.64 &32.67$\pm$0.94 &32.77$\pm$0.67 &33.11$\pm$0.26 &32.68$\pm$0.88 &32.87$\pm$0.59&32.69$\pm$0.88&32.77$\pm$0.86&32.75$\pm$0.16 \\
                     & DP  & 32.08$\pm$0.21 &33.23$\pm$0.15 &32.97$\pm$0.25 &33.24$\pm$0.10 &33.15$\pm$0.16 &33.78$\pm$0.08 &34.08$\pm$0.05 &33.41$\pm$0.07 &33.78$\pm$0.11 &33.45$\pm$0.36 &33.00$\pm$0.30 &34.01$\pm$0.07 &31.99$\pm$0.27&32.83$\pm$0.27&33.64$\pm$0.06&33.24$\pm$0.60\\
                     & SP  & 34.84$\pm$1.67&36.50$\pm$1.41&36.87$\pm$0.48&37.49$\pm$0.37&36.65$\pm$1.52&38.29$\pm$0.40&38.57$\pm$0.34&36.43$\pm$0.61&37.56$\pm$1.30&34.52$\pm$1.61&35.71$\pm$1.78&34.76$\pm$0.34&35.54$\pm$0.28&35.06$\pm$1.65&37.61$\pm$0.40&36.43$\pm$1.27\\
                     & MP  & 32.31$\pm$0.20&32.32$\pm$0.15&33.03$\pm$0.15&32.14$\pm$0.27&33.29$\pm$0.60&34.02$\pm$0.11&33.74$\pm$0.34&34.12$\pm$0.22&33.03$\pm$0.42&32.86$\pm$0.25&32.18$\pm$0.17&34.59$\pm$0.27&32.65$\pm$0.16&32.82$\pm$0.21&33.35$\pm$0.06&33.10$\pm$0.73\\\hline

\multirow{4}{*}{2}   & FT  & 33.16$\pm$0.56 &33.35$\pm$1.13 &33.82$\pm$0.70 &33.24$\pm$0.82 &33.43$\pm$1.12 &33.31$\pm$0.91 &33.30$\pm$1.18 &33.24$\pm$0.69 &33.29$\pm$1.12 &33.19$\pm$0.84 &33.40$\pm$0.57 &33.04$\pm$0.87 &33.20$\pm$0.53&33.03$\pm$0.62&33.29$\pm$0.60&33.29$\pm$0.18\\
                     & DP  & 32.90$\pm$0.74 &35.11$\pm$0.53 &34.44$\pm$0.38 &34.69$\pm$0.51 &35.41$\pm$0.55 &35.43$\pm$0.65 &34.77$\pm$0.55 &34.11$\pm$0.60 &34.93$\pm$0.51 &32.97$\pm$0.37 &35.43$\pm$0.35 &35.19$\pm$0.59 &32.75$\pm$1.22&33.28$\pm$0.94&36.46$\pm$0.65&34.52$\pm$1.07\\
                     & SP  & 35.91$\pm$2.09&38.08$\pm$1.41&38.15$\pm$0.85&38.42$\pm$0.99&37.97$\pm$1.42&38.23$\pm$0.53&38.62$\pm$0.52&36.32$\pm$0.47&39.22$\pm$1.43&34.35$\pm$0.44&37.20$\pm$1.56&34.75$\pm$0.40&35.52$\pm$0.66&36.67$\pm$1.64&37.71$\pm$0.74&37.14$\pm$1.43\\
                     & MP  & 32.76$\pm$0.21&34.25$\pm$0.40&34.10$\pm$0.69&33.26$\pm$0.55&34.59$\pm$0.50&33.81$\pm$0.37&34.33$\pm$0.44&33.75$\pm$0.23&34.01$\pm$0.38&33.88$\pm$0.36&34.55$\pm$0.78&34.51$\pm$0.10&32.59$\pm$0.60&33.83$\pm$0.34&35.39$\pm$1.06&33.97$\pm$0.69 \\\hline

\multirow{4}{*}{4}   & FT  & 33.86$\pm$1.15 &33.89$\pm$1.34 &33.73$\pm$1.12 &33.63$\pm$0.67 &33.90$\pm$1.22 &33.58$\pm$0.94 &33.55$\pm$1.05 &33.86$\pm$1.48 &33.58$\pm$1.27 &33.75$\pm$1.11 &33.71$\pm$1.28 &33.79$\pm$1.43 &33.67$\pm$0.98&33.85$\pm$1.53&33.78$\pm$1.54&33.74$\pm$0.12 \\
                     & DP  & 35.42$\pm$0.46 &37.64$\pm$0.39 &38.85$\pm$0.50 &37.67$\pm$0.53 &39.50$\pm$0.37 &38.91$\pm$0.44 &38.26$\pm$0.23 &36.43$\pm$0.38 &37.54$\pm$0.30 &34.72$\pm$0.24 &37.76$\pm$0.56 &37.23$\pm$0.33 &35.92$\pm$0.53&36.02$\pm$0.56&38.74$\pm$0.40&37.37$\pm$1.36\\
                     & SP  & 38.04$\pm$1.81&40.46$\pm$1.67&40.08$\pm$1.23&40.79$\pm$1.42&41.84$\pm$2.10&39.78$\pm$1.35&41.10$\pm$2.03&37.55$\pm$1.05&41.72$\pm$1.99&35.81$\pm$1.49&39.23$\pm$1.27&35.88$\pm$1.30&37.66$\pm$1.49&37.86$\pm$1.56&39.48$\pm$1.55&39.15$\pm$1.88\\
                     & MP  & 33.14$\pm$0.49&33.79$\pm$1.26&35.16$\pm$1.93&33.95$\pm$1.17&36.26$\pm$2.38&35.52$\pm$0.95&35.44$\pm$1.23&34.63$\pm$0.73&34.21$\pm$1.10&33.53$\pm$0.74&35.96$\pm$1.36&35.62$\pm$1.19&33.51$\pm$0.73&34.06$\pm$0.89&37.10$\pm$0.77&34.79$\pm$1.13\\\hline

\multirow{4}{*}{8}   & FT  & 32.85$\pm$0.44 &32.75$\pm$0.42 &33.05$\pm$0.46 &32.59$\pm$0.54 &33.06$\pm$0.82 &32.58$\pm$0.39 &32.80$\pm$0.38 &32.89$\pm$0.34 &32.88$\pm$0.37 &32.75$\pm$0.24 &33.14$\pm$0.38 &32.69$\pm$0.66 &33.05$\pm$0.27&32.83$\pm$0.41&32.65$\pm$0.31&32.84$\pm$0.17 \\
                     & DP  & 32.73$\pm$0.81 &34.78$\pm$0.63 &34.79$\pm$0.56 &34.82$\pm$0.66 &36.39$\pm$0.43 &34.97$\pm$0.90 &35.17$\pm$0.29 &33.00$\pm$0.50 &34.59$\pm$0.43 &32.91$\pm$0.25 &35.14$\pm$0.60 &34.13$\pm$0.60 &33.14$\pm$0.54&33.66$\pm$0.82&35.56$\pm$0.51&34.39$\pm$1.05\\
                     & SP  & 36.30$\pm$0.94&38.84$\pm$0.68&38.22$\pm$0.34&38.68$\pm$0.69&39.02$\pm$0.57&38.16$\pm$0.86&38.82$\pm$1.08&35.86$\pm$0.53&39.73$\pm$0.45&34.50$\pm$0.69&37.90$\pm$0.71&35.11$\pm$1.05&35.61$\pm$1.22&37.41$\pm$0.98&37.17$\pm$0.64&37.42$\pm$1.54\\
                     & MP  & 32.67$\pm$0.70&33.24$\pm$0.67&34.81$\pm$1.45&33.18$\pm$1.13&34.78$\pm$2.22&34.66$\pm$1.05&34.77$\pm$0.60&34.76$\pm$0.94&33.81$\pm$0.37&33.07$\pm$0.21&34.46$\pm$0.82&35.12$\pm$0.41&32.69$\pm$0.76&33.57$\pm$0.50&36.34$\pm$0.82&34.13$\pm$1.01 \\\hline

\multirow{4}{*}{16}  & FT  & 33.72$\pm$0.84 &34.09$\pm$0.74 &34.28$\pm$0.66 &33.49$\pm$1.15 &34.73$\pm$0.97 &33.82$\pm$0.88 &33.81$\pm$1.13 &33.08$\pm$0.38 &34.06$\pm$1.23 &33.69$\pm$0.88 &33.06$\pm$0.78 &33.57$\pm$0.93 &33.22$\pm$0.45&34.01$\pm$0.63&33.46$\pm$0.91&33.74$\pm$0.44 \\
                     & DP  & 35.07$\pm$1.55 &37.07$\pm$2.07 &37.51$\pm$1.91 &37.43$\pm$2.39 &38.24$\pm$2.15 &36.91$\pm$1.67 &36.61$\pm$1.26 &35.85$\pm$1.67 &36.51$\pm$1.89 &33.84$\pm$0.78 &37.21$\pm$1.45 &35.74$\pm$1.12 &34.86$\pm$1.75&35.77$\pm$1.94&37.86$\pm$1.44&36.43$\pm$1.18\\
                     & SP  & 38.88$\pm$0.92&40.60$\pm$0.43&40.21$\pm$0.66&40.44$\pm$0.87&39.45$\pm$0.71&39.37$\pm$0.55&40.90$\pm$0.43&36.86$\pm$0.59&40.61$\pm$0.53&37.11$\pm$0.61&39.45$\pm$0.43&36.26$\pm$1.04&35.88$\pm$0.47&38.46$\pm$0.70&37.35$\pm$0.95&38.79$\pm$1.65\\
                     & MP  & 32.46$\pm$0.19&33.02$\pm$0.43&33.98$\pm$1.01&32.59$\pm$0.34&33.20$\pm$0.62&34.54$\pm$0.59&34.39$\pm$0.62&34.30$\pm$0.19&33.90$\pm$0.51&33.28$\pm$0.56&33.47$\pm$1.57&34.69$\pm$0.40&32.67$\pm$0.42&33.28$\pm$0.34&35.68$\pm$1.87&33.70$\pm$0.88\\\hline

\multirow{4}{*}{32}  & FT  & 35.84$\pm$1.36 &36.28$\pm$1.47 &36.00$\pm$0.95 &36.11$\pm$1.32 &36.64$\pm$1.46 &36.02$\pm$1.49 &36.47$\pm$1.27 &35.41$\pm$1.26 &35.68$\pm$1.03 &35.33$\pm$1.27 &35.71$\pm$0.98 &35.90$\pm$1.31 &34.81$\pm$1.24&36.10$\pm$1.59&36.20$\pm$1.62&35.90$\pm$0.45    \\
                     & DP  & 41.80$\pm$0.85 &43.51$\pm$0.94 &43.49$\pm$0.72 &42.50$\pm$0.37 &43.65$\pm$0.46 &42.83$\pm$0.66 &43.90$\pm$0.79 &39.30$\pm$1.68 &42.39$\pm$0.83 &37.51$\pm$0.75 &40.51$\pm$1.02 &42.01$\pm$0.76 &39.77$\pm$1.18&41.91$\pm$0.60&39.94$\pm$1.57&41.67$\pm$1.81\\
                     & SP  & 40.30$\pm$1.73&43.38$\pm$0.52&42.08$\pm$0.49&42.27$\pm$1.00&44.72$\pm$0.90&42.32$\pm$0.78&42.34$\pm$0.77&38.91$\pm$1.53&43.76$\pm$0.74&37.54$\pm$1.18&39.97$\pm$1.20&38.79$\pm$0.75&38.83$\pm$0.45&42.09$\pm$1.27&39.56$\pm$1.12 &41.12$\pm$2.06\\
                     & MP  & 40.95$\pm$1.18&42.16$\pm$0.97&42.61$\pm$1.03&42.31$\pm$0.70&45.52$\pm$0.56&41.22$\pm$0.62&44.67$\pm$1.23&40.17$\pm$0.68&42.18$\pm$0.85&36.52$\pm$0.92&40.16$\pm$1.17&41.21$\pm$0.95&40.48$\pm$0.51&41.74$\pm$0.83&40.89$\pm$0.95&41.52$\pm$1.99\\\hline

\multirow{4}{*}{64}  & FT  & 40.16$\pm$2.05 &39.56$\pm$1.83 &40.10$\pm$2.42 &39.87$\pm$1.67 &41.68$\pm$2.26 &40.34$\pm$2.60 &39.47$\pm$2.01 &39.53$\pm$2.11 &38.34$\pm$1.68 &39.64$\pm$1.60 &39.18$\pm$1.81 &39.50$\pm$1.91 &39.23$\pm$1.87&40.85$\pm$2.39&39.63$\pm$2.12&39.81$\pm$0.75 \\
                     & DP  & 45.64$\pm$0.59 &47.64$\pm$0.81 &48.05$\pm$0.66 &46.94$\pm$0.51 &48.89$\pm$0.95 &44.95$\pm$0.85 &47.97$\pm$0.68 &41.61$\pm$0.77 &44.85$\pm$0.61 &40.98$\pm$1.14 &45.65$\pm$0.84 &45.67$\pm$1.15 &43.37$\pm$0.49&47.30$\pm$0.72&45.24$\pm$1.17&45.65$\pm$2.23\\
                     & SP  & 43.48$\pm$0.71&43.81$\pm$0.92&45.99$\pm$0.56&43.70$\pm$0.22&49.04$\pm$0.46&45.79$\pm$0.91&46.11$\pm$0.86&40.86$\pm$0.81&44.51$\pm$1.26&40.49$\pm$0.35&44.68$\pm$0.34&41.91$\pm$1.50&40.09$\pm$0.64&45.25$\pm$0.72&44.17$\pm$0.86&43.99$\pm$2.33\\
                     & MP  & 43.86$\pm$1.05&46.01$\pm$1.13&48.22$\pm$1.20&46.79$\pm$1.39&51.84$\pm$1.12&46.61$\pm$1.39&48.31$\pm$0.86&40.11$\pm$1.23&44.75$\pm$1.24&37.84$\pm$0.53&45.01$\pm$1.02&44.82$\pm$1.22&43.95$\pm$1.25&48.28$\pm$1.16&43.03$\pm$1.50&45.30$\pm$3.33\\\hline

\multirow{4}{*}{128} & FT  & 43.50$\pm$1.77 &45.52$\pm$2.29 &45.60$\pm$2.40 &44.38$\pm$2.56 &46.94$\pm$2.98 &45.75$\pm$2.75 &46.00$\pm$1.91 &42.96$\pm$2.64 &44.94$\pm$2.28 &41.43$\pm$1.88 &43.27$\pm$2.02 &43.67$\pm$2.56 &41.78$\pm$2.46&44.81$\pm$2.44&44.79$\pm$1.67&44.36$\pm$1.52 \\
                     & DP  & 46.23$\pm$1.06 &50.49$\pm$0.84 &50.99$\pm$0.75 &47.39$\pm$0.80 &53.68$\pm$0.48 &48.53$\pm$1.00 &49.28$\pm$0.56 &44.77$\pm$0.95 &46.93$\pm$0.54 &42.03$\pm$0.68 &47.95$\pm$0.66 &49.56$\pm$0.71 &44.21$\pm$0.84&48.92$\pm$1.36&49.56$\pm$0.77&48.03$\pm$2.83\\
                     & SP  & 44.78$\pm$4.43&46.24$\pm$3.94&45.30$\pm$4.21&46.31$\pm$3.70&49.45$\pm$5.63&45.80$\pm$4.61&46.37$\pm$4.48&43.29$\pm$4.19&44.95$\pm$3.74&41.21$\pm$3.54&45.64$\pm$3.95&41.93$\pm$4.95&41.18$\pm$4.09&44.99$\pm$4.54&45.73$\pm$4.93&44.88$\pm$2.13\\
                     & MP  & 46.48$\pm$2.40&47.98$\pm$1.98&49.04$\pm$1.99&49.09$\pm$1.32&52.55$\pm$0.86&49.66$\pm$1.52&50.34$\pm$1.28&47.03$\pm$1.40&46.40$\pm$2.25&42.89$\pm$1.95&48.08$\pm$1.65&48.45$\pm$1.25&44.04$\pm$1.99&48.15$\pm$2.19&50.47$\pm$1.55&48.04$\pm$2.38\\\hline

\multirow{3}{*}{256} & FT  & 52.13$\pm$2.07 &54.57$\pm$1.95 &54.43$\pm$2.26 &54.00$\pm$1.63 &57.79$\pm$2.36 &55.89$\pm$2.37 &55.39$\pm$1.97 &50.65$\pm$1.71 &52.90$\pm$2.03 &50.00$\pm$1.89 &51.22$\pm$1.92 &52.31$\pm$2.06 &48.57$\pm$2.25&54.16$\pm$1.71&52.10$\pm$2.60&53.07$\pm$2.35\\
                     & DP  & 53.23$\pm$0.56 &55.59$\pm$1.40 &55.39$\pm$1.43 &55.05$\pm$0.86 &60.14$\pm$0.67 &50.64$\pm$0.83 &54.43$\pm$1.29 &46.10$\pm$1.77 &51.35$\pm$0.86 &45.26$\pm$1.57 &53.42$\pm$0.86 &50.83$\pm$0.71 &48.42$\pm$1.10&55.14$\pm$0.72&52.72$\pm$0.68&52.51$\pm$3.76\\
                     & SP  & 52.26$\pm$1.89&56.04$\pm$1.71&53.02$\pm$2.14&53.12$\pm$1.62&60.58$\pm$0.86&54.80$\pm$1.42&55.79$\pm$1.15&49.43$\pm$1.87&52.49$\pm$1.78&47.33$\pm$1.11&54.52$\pm$1.21&52.08$\pm$1.49&48.48$\pm$0.93&54.54$\pm$1.05&54.59$\pm$1.31&53.27$\pm$3.17\\
                     & MP  & 52.77$\pm$0.31&53.98$\pm$1.26&50.71$\pm$2.41&54.63$\pm$0.95&60.13$\pm$0.40&51.64$\pm$2.83&55.32$\pm$0.79&49.58$\pm$1.87&53.50$\pm$0.68&45.27$\pm$0.70&53.37$\pm$0.23&51.28$\pm$1.26&47.16$\pm$0.75&52.34$\pm$1.15&53.80$\pm$0.82&52.37$\pm$3.38\\\hline
\end{tabular}
\caption{Zero-shot crosslingual transfer results in accuracy
  (\%). We report mean and variance of 5 runs, when using finetuning (FT),
  discrete prompting (DP), soft prompting (SP), and mixed prompting (MP). ``MAJ'':
  majority baseline; $\overline{X}$: macro average across 15
  languages.}
\tablabel{zeroshottablewithvariance}
\end{table}


\end{landscape}


\end{document}